\documentclass[10pt,final,3p]{elsarticle}
\usepackage{lmodern}
\usepackage{soul}

\usepackage{graphics}
\usepackage{graphicx}

\usepackage{booktabs}
\usepackage{multirow}
\usepackage{array}
\usepackage{amsmath}

\usepackage{amsthm}
\usepackage{verbatim}
\usepackage{relsize}
\usepackage[dvipsnames]{xcolor}
\usepackage{array}
\usepackage{bm}

\usepackage{caption}

\usepackage{algorithm}
\usepackage{algpseudocode}

\usepackage{natbib}
\usepackage{hyperref}
\newcommand{\doi}[1]{\textsc{doi}: \href{http://dx.doi.org/#1}{\nolinkurl{#1}}}

\definecolor{comment}{rgb}{0.0, 0.5, 0.0}

\hypersetup{
    colorlinks=true,
    linkcolor=blue,
    filecolor=magenta,
    urlcolor=cyan,
}
\usepackage{booktabs}
\usepackage{xcolor}
\usepackage{color}

\definecolor{mygreen}{rgb}{0,0.6,0}
\definecolor{mygray}{rgb}{0.5,0.5,0.5}
\definecolor{mymauve}{rgb}{0.58,0,0.82}

\usepackage{array}
\usepackage{mathtools}
\usepackage{verbatim}
\usepackage{relsize}
\usepackage{multirow}
\usepackage{multicol}

\newcommand{\lstin}[1]{{\ttfamily #1}}
\usepackage{listings}
\usepackage[bitstream-charter]{mathdesign}
\usepackage[T1]{fontenc}
\usepackage[utf8]{inputenc} 

\usepackage[english]{babel}
\usepackage[normalem]{ulem}

\DeclareMathAlphabet{\pazocal}{OMS}{zplm}{m}{n}

\usepackage{lscape}		
\usepackage{fancyvrb}

\usepackage[mathscr]{euscript}

\usepackage{textcomp}

\def \balpha {{\bm{\alpha}}}
\def \bbeta  {{\bm{\beta}}}

\def \bPsi {{\boldsymbol\Psi}}
\def \xxi  {{\boldsymbol{\xi}}}

\def \X  {\bm{X}} 
\def \x  {\bm{x}} 

\def \cand {\mathrm{c}}
\def \neigh {\mathrm{s}}

\usepackage{bigints}

\usepackage[switch]{lineno}

\newcounter{MVremark}

  \usepackage{lineno}

\journal{Engineering Structures}

\begin{document}

\begin{frontmatter}



\title{Uncertainty Quantification of Engineering Structures by Polynomial Chaos Expansion and Multivariate Active Learning }

\author{Qitian Lu}  \ead{Qitian.Lu@vutbr.cz}
\address{Brno University of Technology, Brno, Czech Republic}

\author{Jafar Jafari-Asl} \ead{jafar.jafariasl@uni-rostock.de}
\author{Panagiotis Spyridis} \ead{panagiotis.spyridis@uni-rostock.de} 
\address{University of Rostock, Rostock, Germany}

\author{Luk{\'a}{\v s} Nov{\'a}k\texorpdfstring{\corref{cor1}}{}}  \ead{Lukas.Novak4@vut.cz}  \cortext[cor1]{Corresponding author}
\address{Brno University of Technology, Brno, Czech Republic}

\begin{abstract}
In many engineering applications, a single high-fidelity model produces multiple quantities of interest (QoIs) under the same input parameters, e.g. finite element models of complex physical systems. To alleviate the high computational cost of direct model evaluations, surrogate models are widely used to construct efficient approximations of model responses. Naturally, the accuracy of surrogates strongly depends on the quality of the experimental design (ED). However, a single ED may not provide an adequate representation for all outputs simultaneously, especially when different outputs exhibit varying sensitivities to the input variables. A straightforward solution is to perform separate sampling for each output, but this results in increased sampling complexity and computational cost. From a statistical perspective, such an approach also ignores potential correlations among all outputs and may compromise data consistency. To address this issue, an adaptive sequential sampling method for constructing polynomial chaos expansion surrogate models is generalized for vector valued QoIs. The method sequentially selects new samples from a candidate pool based on their local contribution to the output variance, while balancing distance-based exploration of the input space and exploitation of aggregated variance information across all outputs. Its performance is compared with non-sequential Latin Hypercube Sampling through several numerical examples from engineering problems. Numerical results demonstrate that the proposed strategy improves both surrogate accuracy and stability, and provides a more reliable estimation of second-order statistics.
\end{abstract}


\begin{highlights}
\item Generalization and application of multi-output sequential adaptive sampling are presented.
\item A normalized variance aggregation is introduced to evaluate multi-output PCE models.
\item The utilized criterion balances variance contribution and spatial exploration of the samples for multi-output problems.
\item Numerical results of engineering examples demonstrate superior accuracy and stability of the active learning.

\end{highlights}


\end{frontmatter}


\section{Introduction}
\label{Introduction}

Uncertainty is an inseparable part of all engineering systems and can affect structural behavior in unpredictable ways. This phenomenon arises from numerous factors such as environmental conditions and construction errors, which influence the applied loads, material properties, and the dimensions of structural members.  Therefore, by determining the precise value of these uncertainties, an accurate estimation of structural performance and behavior over their lifetime can be obtained. This estimation includes the safety level and the most probable failure modes, allowing for necessary actions to be taken on time when required. The main goal is thus to determine the amount, understand the impact, describe, and propagate the uncertainties present in engineering structures using mathematical models. In the real world, uncertainty quantification (UQ) for engineering structures requires complex simulations to model structural responses using finite element based models, which usually have a very high computational cost. As a result, using numerical and traditional Monte Carlo (MC) simulation methods is often not feasible due to very high computational burden. Therefore, the use of surrogate models for UQ in engineering structures has attracted significant attention.

Moreover, in structural and mechanical analysis, a high-fidelity numerical simulation often provides not one but several quantities of interest simultaneously. This multi-output nature becomes particularly challenging in uncertainty-related analyses, where repeated model evaluations are required and the associated computational cost may become prohibitive. Surrogate modeling therefore provides an effective way to alleviate this burden while maintaining acceptable prediction accuracy. Among the available surrogate modeling techniques, polynomial chaos expansion (PCE) \cite{Wiener_PCE} has become a well-established approach for UQ \cite{LuthenReview}, sensitivity analysis \cite{SUDRET,CRESTAUX20091161}, reliability analysis \cite{Zhou2020, Marelli2018}, and recently scientific machine learning \cite{NOVAK2024112926, SHARMA2024117314} or operator learning \cite{SHARMA2026118796}. Owing to its natural mathematical structure, PCE provides not only an efficient approximation of random system responses, but also direct access to statistical quantities \cite{NOVAK2022106808}, such as moments, and sensitivity indicators through the expansion coefficients \cite{PCEMoments}. This property makes non-intrusive PCE particularly attractive for applications with expensive numerical models, where the total number of admissible model evaluations is strongly limited, e. g. complex non-linear finite element model of a prestressed concrete bridge presented by Slowik et al. \cite{ApplicationPCE_bridge}. 

However, for non-intrusive PCE constructed by regression, the quality of the surrogate highly depends on the ED \cite{PCEOLSreview}, i.e., quality of the training data set. Various work has been devoted to the design of sampling strategies for surrogate construction. Classical static sampling designs include MC sampling \cite{McKayConovBeck:three:1979}, Latin hypercube sampling (LHS) \cite{Stein:87:LargeSampleLHS}, and quasi-Monte Carlo methods \cite{Nieder:RandNumGen_AND_QMC_1992} based on low-discrepancy sequences. In addition, from a geometric and distributional perspective, space-filling principles have led to a variety of sampling criteria, such as miniMax, Maximin \cite{JohMooYlv:MixiMinMiniMax:JSPI:90}, generalized versions of distance-based criteria \cite{MorMit:JStatPLanInf:95,VorEli:Technometrics:20}, discrepancy-based criteria \cite{fang2001wrap}, and entropy-based criteria \cite{Shewry1987MaximumES}. From a view of regression, several optimality criteria, including D-optimality, A-optimality, E-optimality, and I-optimality \cite{AtkinsonDonev1992,Duarte2025}, have also been employed to improve the stability and accuracy of surrogate model by optimizing various characteristics of information matrix. These approaches are attractive because of their ease of implementation. However, since all design points are selected a priori at once, these strategies cannot exploit the information extracted from existing evaluated samples. Moreover, their efficiency may deteriorate when the response becomes complex or non-uniform, especially in multi-output settings where different outputs may show different levels of complexity across the input space. In order to make full use of the information provided by the existing surrogate and available responses, adaptive sequential strategies, often referenced as active learning techniques, have been investigated in surrogate modeling and reliability analysis \cite{Zhou2020}. In these methods, the ED is iteratively refined by selecting new points according to the information provided by the continuously updated surrogate model. The criteria are often based on the approximation error, statistical indicator, or local geometry of the input space. For PCE, adaptive sequential design is typically guided by criteria tailored to the structure of the expansion \cite{AdaptiveCoherence,SequentialPCEThapa}. In practice, the criteria often rely on error indicators, basis adaptivity, or variance-related measures that exploit the information contained in the PCE coefficients \cite{PCESoptimSeq,NOVAK2021114105}. Within this context, the variance-based adaptive sequential sampling strategy proposed by Nov\'ak et al. represents an important development for PCE-based surrogate construction, where the associated $\Theta$ criterion combines a variance-related term with a distance-based term to achieve a balance between exploitation and exploration \cite{NOVAK2021114105}. This balance has proved particularly appealing in engineering applications \cite{Theta_application1,ThetaApplication_2} because it avoids both excessive concentration of samples and poor coverage in parts of the input space, while favoring areas that are expected to contribute significantly to the total response variance. As a result, the $\Theta$ criterion has shown promising performance for the adaptive construction of PCE surrogates.

Despite these merits, the original $\Theta$ criterion is formulated for single-output problems. This restriction may be limiting in many engineering applications, where one simulation simultaneously produces several outputs, and all of them must be approximated under a limited and shared computational budget. The straightforward single-output criterion is generally unsatisfactory for the following reasons. First, using different ED for different outputs may make it difficult to maintain consistency across the surrogate construction, which is undesirable when multiple responses are to be approximated jointly in a unified multi-output setting. Second, the repeated sampling procedure for different training sets may become computationally inefficient when the underlying model is expensive. Third, different outputs may exhibit different levels of variability, smoothness, and approximation difficulty, so that a sample selected as informative for one output may be much less informative for other outputs. Therefore, the extension from single-output to multi-output criterion is not merely a formal generalization but a necessary step toward an efficient sampling strategy for practical multi-output problems.

Motivated by this limitation, this paper proposes a multi-output $\Theta$ criterion for the adaptive sequential sampling for non-intrusive PCE. The proposed approach aggregates variance and distance information from multiple outputs into a unified criterion, so that each additional sample is selected to maximize the overall improvement of the surrogate model. The proposed method is verified through several numerical examples, including engineering problems, and is compared to LHS, as a standard sampling method for practical applications, in terms of approximation accuracy, statistical estimation performance, and model stability. The results show that the proposed method provides an effective and efficient sampling strategy for multi-output PCE surrogate models.

\section{Non-intrusive Polynomial Chaos Expansion}
\label{PCE section}
Let $ Y=g\left(\X\right)$ denote a mathematical model, where $\X=(X_1, X_2,\dots, X_M)$ represents a set of $M$ independent random input variables. In the framework of standard PCE, the input variables are first mapped to a standardized random germ $\xxi=(\xi_1, \xi_2,...,\xi_M)$, which enables construction of orthogonal polynomial bases according to the Wiener–Askey scheme \cite{Askey}. The orthonormal condition is defined as
\begin{equation}
\langle \psi_j, \psi_k \rangle =
\int \psi_j(\xi)\psi_k(\xi)p_{\xi}(\xi)d\xi =0, \quad j\neq k.
\end{equation}
where $\langle \cdot,\cdot \rangle$ denotes the inner product, $\psi_j$ and $\psi_k$ are orthogonal polynomial basis functions, and $p_{\xi}(\xi)$ is the probability density function of the germ $\xi$. The corresponding multivariate polynomial basis function $ \Psi (\xxi)$ is constructed as a tensor product of univariate orthonormal polynomials
\begin{equation}
\label{Eq:MultVarPol}
    \Psi_{\balpha} \left( \xxi \right)
    =
    \prod_{i=1}^{M}  \psi_{\alpha_i} \left(\xi_i\right),
\end{equation}
where $ {\balpha}\in \mathbb{N}^M $ is a multi-index indicating the polynomial degree associated with each dimension $\xi_i$. Accordingly, the response $Y$ can be further represented using these multivariate basis functions as ~\cite{ghanem2003stochastic}
\begin{equation}
\label{PCE}
    Y  = g\left(\X\right) =
    \sum_{\balpha \in \mathbb{N}^M }
    \beta_{\balpha}\Psi_{\balpha}\left( \xxi\right),
\end{equation}
where $\beta_{\balpha}$ are the deterministic coefficients of multivariate polynomial basis. These coefficients can be estimated by minimizing the approximation error using regression techniques such as ordinary least squares (OLS) or least angle regression (LAR).

For practical computation, the PCE model must be truncated to a finite number of terms. A commonly used truncation strategy is the total-degree truncation, where only multivariate polynomial terms with total degree not exceeding a prescribed order $p$ will be retained. Then, the resulting number of basis terms is given by
\begin{equation}
P = \frac{(M+p)!}{M!\,p!}.
\end{equation}
As $M$ or $p$ increases, the number of basis terms $P$ grows rapidly, which leads to a significant increase in computational cost - commonly known phenomenon referenced as a curse of dimensionality. Therefore, an appropriate application of the basis selection strategy becomes crucial. The OLS is commonly used when the number of basis functions is moderate, LAR and other advanced basis selection algorithms~\cite{OMP, BCS} are often adopted for high-dimensional sparse problems where the number of candidate basis functions becomes far larger than necessary. In the case of OLS, the deterministic coefficients $\bbeta$ can be easily estimated based on the ED, expressed as
\begin{equation}
    \bbeta
    =
    \left(\bPsi^{T}\bPsi\right)^{-1} \ \bPsi^{T}  \mathbf{y},
\label{Eq:PCe_OLS}
\end{equation}
where $\mathbf{y}$ represents the output of the training set, and $ \bPsi $ is the data matrix constructed by evaluating the basis functions at the input samples in the ED.

Once a PCE surrogate model is established, its predictive capability must be quantified. For this purpose, the leave-one-out (LOO) cross-validation error $Q^2$ is commonly employed, which is less prone to overfitting compared to the coefficient of determination $R^2$. The LOO error is obtained by evaluating the prediction error when each sample in the ED is excluded once from the model construction. For each exclusion, a surrogate model is built using the remaining samples and the prediction error is computed at the omitted point. Repeating this procedure for all samples provides an estimate of the overall prediction error. Although a direct implementation of this procedure would require a large computational cost, the LOO error $Q^2$ can also be computed analytically from a single PCE model as follows:
\begin{equation}
Q^2 =
1 -
\frac{
\frac{1}{n_{\mathrm{sim}}}
\sum_{i=1}^{n_{\mathrm{sim}}}
\left[
\frac{g\!\left(x^{(i)}\right) - g^{\mathrm{PCE}}\!\left(x^{(i)}\right)}
{1-h_i}
\right]^2
}
{\sigma_{Y,\mathrm{ED}}^{2}}
\end{equation}
where $\sigma_{Y,\mathrm{ED}}^{2}$ is the variance of the model responses evaluated at the ED points using the original model and $h_i$ denotes the $i$th diagonal term of the matrix $\mathbf{H} = \mathbf{\Psi}\left(\mathbf{\Psi}^{T}\mathbf{\Psi}\right)^{-1}\mathbf{\Psi}^{T}.
$

Moreover, the specific structure of the PCE facilitates the calculation of the statistical moments of the response. The raw $m$th order statistical moment can be rewritten in terms of the PCE as follows:
\begin{equation}
\begin{aligned}
\langle Y^{m} \rangle
&= \int [g(\mathbf{X})]^m p_{\mathbf{X}}(\mathbf{X})\, d\mathbf{X}  \\
&= \int \left[ \sum_{\alpha \in \mathbb{N}^M}
\beta_{\alpha}\Psi_{\alpha}(\boldsymbol{\xi}) \right]^m
p_{\boldsymbol{\xi}}(\boldsymbol{\xi})\, d\boldsymbol{\xi}  \\
&= \int
\sum_{\alpha_1\in\mathbb{N}^M}\cdots
\sum_{\alpha_m\in\mathbb{N}^M}
\beta_{\alpha_1}\cdots\beta_{\alpha_m}
\Psi_{\alpha_1}(\boldsymbol{\xi})\cdots
\Psi_{\alpha_m}(\boldsymbol{\xi})
p_{\boldsymbol{\xi}}(\boldsymbol{\xi})\, d\boldsymbol{\xi} \\
&=
\sum_{\alpha_1\in\mathbb{N}^M}\cdots
\sum_{\alpha_m\in\mathbb{N}^M}
\beta_{\alpha_1}\cdots\beta_{\alpha_m}
\int
\Psi_{\alpha_1}(\boldsymbol{\xi})\cdots
\Psi_{\alpha_m}(\boldsymbol{\xi})
p_{\boldsymbol{\xi}}(\boldsymbol{\xi})\, d\boldsymbol{\xi}.
\end{aligned}
\label{Eq:PCE_moment}
\end{equation}
As an alternative to integrating the original model $g(\mathbf{X})$, the integration now is performed only on the basis functions, greatly simplifying the computation. Specifically, due to the orthonormality of the basis functions, the first and second statistical moments can be further simplified as
\begin{equation}
\langle Y^{1} \rangle =\mu_Y =  \beta_0
\end{equation}
and
\begin{equation}
\langle Y^{2} \rangle
=
\sum_{\alpha \in \mathcal{A}}
\beta_{\alpha}^{2}
\langle \Psi_{\alpha}, \Psi_{\alpha} \rangle .
\end{equation}
Then, the variance of the response $\sigma_Y^2 = \langle Y^{2} \rangle - \mu_Y^2$ can be readily obtained from these two moments as
\begin{equation}
\sigma_Y^2 =
\sum_{\substack{\alpha \in \mathcal{A} \\ \alpha \neq 0}}
\beta_{\alpha}^{2}.
\end{equation}
Owing to the convenient variance evaluation enabled by the PCE representation, and motivated by the connection between variance-based measures and the overall variation of a function, such as the variation in the sense of Hardy and Krause~\cite{Koksma:ineq_1942}, the response variance is adopted as an important component of the subsequent $\Theta$-criterion sampling strategy. However, for adaptive sampling, a global variance is not sufficient, since it does not indicate where new samples should be placed. It is necessary to identify the regions of the input space that contribute most significantly to the total variance. To this end, the total variance is further localized by interpreting it as an integral of pointwise contributions over the input space. Based on the PCE representation, a local variance measure can then be defined to quantify the contribution of a given point $\boldsymbol{\xi}$ to the total variance, expressed as

\begin{equation}
\sigma_\mathcal{A}^2(\boldsymbol{\xi}) =
\left[
\sum_{\substack{\alpha \in \mathcal{A} \\ \alpha \neq 0}}
\beta_\alpha \, \Psi_\alpha(\boldsymbol{\xi})
\right]^2
\, p_{\boldsymbol{\xi}}(\boldsymbol{\xi}),
\end{equation}
where a large value of $\sigma_\mathcal{A}^2(\boldsymbol{\xi})$ indicates that the corresponding region contributes more to the total variance of the surrogate response and is therefore important for adaptive sampling.

\section{Adaptive Sequential Sampling}
\label{AL section}
When the coefficients of a non-intrusive PCE are estimated using regression techniques, the accuracy and stability of the surrogate model depend strongly on the quality of the ED. In many practical engineering problems, the costs associated with the construction of ED, i.e. evaluations of the original numerical model, are often very high. This issue becomes particularly significant among multi-output problems, where samples may contribute unevenly to approximation of different outputs, requiring a larger number of training samples to achieve satisfactory accuracy across all outputs. Therefore, an appropriate sampling strategy is essential for constructing a reliable surrogate model when only a limited number of training samples is available. 

Traditional ED approaches typically rely on static sampling strategies, in which all training points are selected in advance based solely on the input space and the assumed design objective, without exploiting the outputs of previously evaluated samples or structures of existing surrogates \cite{GARUD201771}. Owing to their generality and ease of implementation, static designs have been widely adopted as a common choice for surrogate construction, particularly when prior knowledge of the response is limited. One of the earliest and most often used choice is MC sampling, which is attractive because of its simplicity and its natural consistency with the prescribed probability distribution of the inputs. However, purely random sampling for low to mid-size ED may suffer from poor geometric coverage (clustering and large empty regions in design space) and limited sampling efficiency when the available training budget is small. To mitigate the problem caused by uneven sample distribution, subsequent developments introduce more structured sampling strategies to improve the coverage of the input domain. A classic representative is LHS, which partitions each input dimension into multiple adjacent, non-overlapping subintervals to improve the coverage of the input space and reduce the risk that some regions remain insufficiently explored \cite{Stein:87:LargeSampleLHS}. For this reason, LHS becomes one of the most widely used static sampling methods in surrogate modeling, sensitivity analysis, and UQ, especially for high-dimensional problems. Beyond marginal stratification, another important design idea is to improve the uniformity of samples in the global domain. This consideration motivates the use of Quasi Monte Carlo \cite{Nieder:RandNumGen_AND_QMC_1992} methods based on low discrepancy sequences, such as Sobol, Halton, and Hammersley sequances, which aim to distribute samples more uniformly over the input domain than purely random sampling. 

Another related but explicitly geometric strategy is space-filling design, under which the design problem is no longer regarded merely as sampling from the input distribution, but rather as selecting points that fill the admissible region effectively. This line of development led to various criterial, including miniMax, Maximin \cite{JohMooYlv:MixiMinMiniMax:JSPI:90}, generalized versions of distance-based criteria \cite{MorMit:JStatPLanInf:95,VorEli:Technometrics:20}, discrepancy-based criteria \cite{fang2001wrap}, and entropy-based criteria \cite{Shewry1987MaximumES}.

Importantly, space-filling is one of the possible design principle for static ED. When a surrogate model is constructed through regression, another important consideration is the prediction accuracy, based on which the model-based optimal criteria are proposed. In this class, the ED is selected to improve the conditioning of the regression, thereby enhancing the accuracy and stability of the surrogate model. Representative examples include D-optimality, A-optimality, and E-optimality \cite{Kieferoptimal}, which are defined based on the information of the covariance matrix, whereas I-optimality \cite{Ioptimal} focuses on minimizing the average prediction variance over the design domain.

Despite their differences, all above strategies share a common limitation: the ED is selected before any response is observed. Consequently, they cannot exploit the information revealed by the existing evaluated samples. This issue becomes more critical in multi-output problems, where different outputs may exhibit different levels of complexity, and a sample that is informative for one output may contribute less to the others. As a result, static designs may fail to improve all outputs in a balanced manner, and more training samples are required to ensure satisfactory accuracy of the most complicated output. Therefore, it is desirable to develop an adaptive sequential sampling criterion to fully utilize the information contained in the current training samples. To this end, a multi-output $\Theta$ criterion is developed from the single-output $\Theta$ criterion originally proposed in \cite{NOVAK2021114105} to guide the sequential selection of samples for PCE surrogate.

We propose an adaptive sequential sampling strategy aimed at improving the estimation of the PCE coefficients $\bbeta$ for multi-output problems. The strategy relies on a candidate pool consisting of $n_{\mathrm{pool}} $ realizations of the random input vector $\xxi$ generated by an arbitrary sampling technique. The enrichment process is initialized using a surrogate model constructed by a relatively small initial ED. To account for the spatial distribution of the samples and the variance contribution of the model response, the $\Theta$ criterion is extended to the multi-output case as follows:
\begin{equation}
\label{Eq:ThetaCrit}
\Theta(\xxi^{(\mathrm{c})})
=
\underbrace{
\sqrt{
\left(
\sum_{r=1}^{N_{\mathrm{out}}}
\frac{\sigma_{{r}}^2(\xxi^{(\mathrm{c})})}
{\displaystyle \max_{\xxi_i^{(\mathrm{c})} \in \mathcal{S}_{\mathrm{c}}}
\sigma_{r}^2(\xxi_i^{(\mathrm{c})})}
\right)
\left(
\sum_{r=1}^{N_{\mathrm{out}}}
\frac{\sigma_{r}^2(\xxi^{(\mathrm{s})})}
{\displaystyle \max_{\xxi_i^{(\mathrm{s})} \in \mathcal{S}_{\mathrm{s}}}
\sigma_{r}^2(\xxi_i^{(\mathrm{s})})}
\right)
}
}_{\text{ave normalized  variance aggregation}}
\;
\times l_{\mathrm{c},\mathrm{s}}^{M}
\end{equation}
where $\xxi^{(\mathrm{c})}$ denotes a candidate point from the candidate pool $\mathcal{S}_{\mathrm{c}}$, and $\xxi^{(\mathrm{s})}$ denotes its nearest neighbor in the current ED. $\sigma_{r}^{2}(\xxi)$ denotes the local variance associated with the $r$th ($r=1,\dots,N_{\mathrm{out}}$) output evaluated at point $\xxi$. The normalization factors correspond to the maximum local variance of the $r$th output component evaluated over the candidate set $\mathcal{S}_{\mathrm{c}}$ and over the corresponding nearest-neighbor set $\mathcal{S}_{\mathrm{s}}$, respectively. The distance $l_{\mathrm{c},\mathrm{s}}$ represents the Euclidean distance between the candidate point and its nearest neighbor in the input space, defined as
\begin{equation}
    \label{eq:metric}
    l_{\mathrm{c},\mathrm{s}}=\sqrt{\sum_{i=1}^{M} |\xi_i^{(\cand)}-\xi_i^{(\neigh)}|^2  }.
\end{equation}
where $M$ is the dimension of the input vector. Furthermore, we introduce an abbreviated notation by dropping the explicit dependence on the polynomial basis index set $\mathcal{A}$, which may change during the adaptive construction of the PCE model as the ED is enriched. Multiplication of these two independent contributions maintains the optimal balance between exploitation (denoted as “ave normalized variance aggregation”) and exploration (the distance term $l_{\mathrm{c},\mathrm{s}}$).

The exploitation component of the proposed criterion is motivated by the idea of refining regions that contribute significantly to the overall variance of the model response. For a PCE surrogate, the variance of each output can be interpreted as the integral of local variance contributions over the input domain. This interpretation allows the introduction of a local variance measure that quantifies the contribution of a given location in the input space to the overall model variability. When evaluating a candidate point $\xxi^{(\mathrm{c})}$, it is natural to consider the region between this point and its nearest neighbor $\xxi^{(\mathrm{s})}$ in the current ED. The variance contribution of this region can be approximated by combining the local variance measures at the two points. In particular, the geometric mean of the normalized variance aggregations at the candidate and its nearest neighbor provides an estimate of the average variance level within this region. Multiplying this quantity by the distance term $l_{\mathrm{c},\mathrm{s}}^{M}$ yields an approximation of the variance contained in the neighborhood between the two points. Consequently, the $\Theta$ criterion can be interpreted as an estimate of the amount of unresolved variance that would be refined by selecting the candidate. In this way, the criterion naturally favors candidates located in regions with both high variance contribution and low sampling density.

In the particular case where the variance density is constant over the input domain, the proposed criterion is simplified to a purely space-filling strategy, since the variance aggregation term becomes uniform for all candidates and the selection is driven solely by the distance term $l_{\mathrm{c},\mathrm{s}}^{M}$. In this situation, the criterion behaves similarly to a miniMax design \cite{JohMooYlv:MixiMinMiniMax:JSPI:90,EliVorSad:miniMax:ADES:20}, which promotes a uniform distribution of samples over the design domain. The geometric mean is used instead of the arithmetic mean in order to avoid selecting candidates located extremely far from the region of significant probability mass. In problems with unbounded input domains (e.g., Gaussian variables), candidates far from the mean may exhibit very large distances but negligible variance contribution. The geometric mean naturally penalizes such cases and prevents the selection of candidates with vanishing variance density. Therefore, maximization of the $\Theta$ criterion identifies candidates that are expected to refine regions with high variance contribution and insufficient sampling density, enabling efficient sequential enrichment of the ED.

\section{Numerical Examples and Engineering Applications}
The proposed $\Theta$ criterion is numerically tested on several multi-output examples and compared with LHS. To comprehensively assess the model accuracy, various error measures are employed, including the mean absolute error (MAE), the maximum absolute error $\mathrm{AE}_{\max}$, and the relative variance error $\varepsilon_{var}$ defined as
\begin{eqnarray}
    \begin{aligned}
        &\mathrm{MAE}=\frac{\sum_{i=1}^{n_{\mathrm{test}}}\lvert y_i^{pce}-y_i^{ref}\rvert}{n_{test}},\quad\mathrm{AE}_{\max} = \max_{i=1,\dots,n_{\mathrm{test}}} \left| y_i^{\mathrm{pce}} - y_i^{\mathrm{ref}} \right|, \\
        &\quad\quad\quad\quad\quad\quad\quad\quad\quad\mathrm{\varepsilon_{var}} =\frac{ \lvert \sigma^2-\sigma_y^2 \rvert}{\sigma_y^2}.
    \end{aligned}
    \label{Eq. error}
\end{eqnarray}
To further investigate the stability of the surrogate model, the LOO error is also considered. For a consistent visualization, we use $1-Q^2$ in the following plots. For all examples, the calculations are repeated in order to get statistical information, i.e. the mean values of above error measures as well as the corresponding $\pm \sigma$ intervals.

\subsection{Illustrative Example: 2D Mirror Line Singularities}
To illustrate the performance of the proposed $\Theta$ criterion for sequential sampling, we first consider a 2D line-singularity function exhibiting point symmetry about the center of input space, $(0.5,0.5)$. The mathematical formulation of the model is given by
\begin{equation}
    Y = f(\boldsymbol{X})=\frac{1}{ \lvert 0.3-X_1^2 - X_2^2\rvert + \delta}
    -
    \frac{1}{ \lvert 0.3-(1-X_1)^2 - (1-X_2)^2\rvert + \delta},
    \qquad
    (X_1, X_2) \in [0,1]^2.
\end{equation}
This model contains two terms whose denominators vanish along circular curves in the input domain. Singular behavior localized near these curves induces steep gradients confined to narrow regions, which makes the problem challenging for the PCE approximation, particularly when the parameter $\delta$ is small. In this study, we focus on the case of $ \delta = 0.1 $, for which the variance of the response is approximately $\sigma_Y^2 \approx 13.070 \, 477 \, 042$. Due to the narrow singular regions, space-filling sampling strategies may fail to allocate sufficient samples in these regions. Consequently, a global PCE surrogate model trained on such samples may not adequately capture the localized features. This indicates the potential applicability of a variance-based criterion to guide adaptive sampling toward regions that contribute most significantly to predictive uncertainty. Moreover, for the purpose of constructing a multi-output task, we introduce a diagonal partition $x_1=x_2$ and treat the response in each subregion as a distinct output component, thereby reformulating the original single-output problem into a multi-output task:

\begin{equation}
Y_1(\boldsymbol{X})=
\begin{cases}
f(\boldsymbol{X}), & X_1\ge X_2,\\
0, & X_1< X_2,
\end{cases}
\qquad
Y_2(\boldsymbol{X})=
\begin{cases}
0, & X_1>  X_2,\\
f(\boldsymbol{X}), & X_1\leq X_2.
\end{cases}
\end{equation}

\begin{figure}[t]
    \centering
    \includegraphics[width=0.8\linewidth]{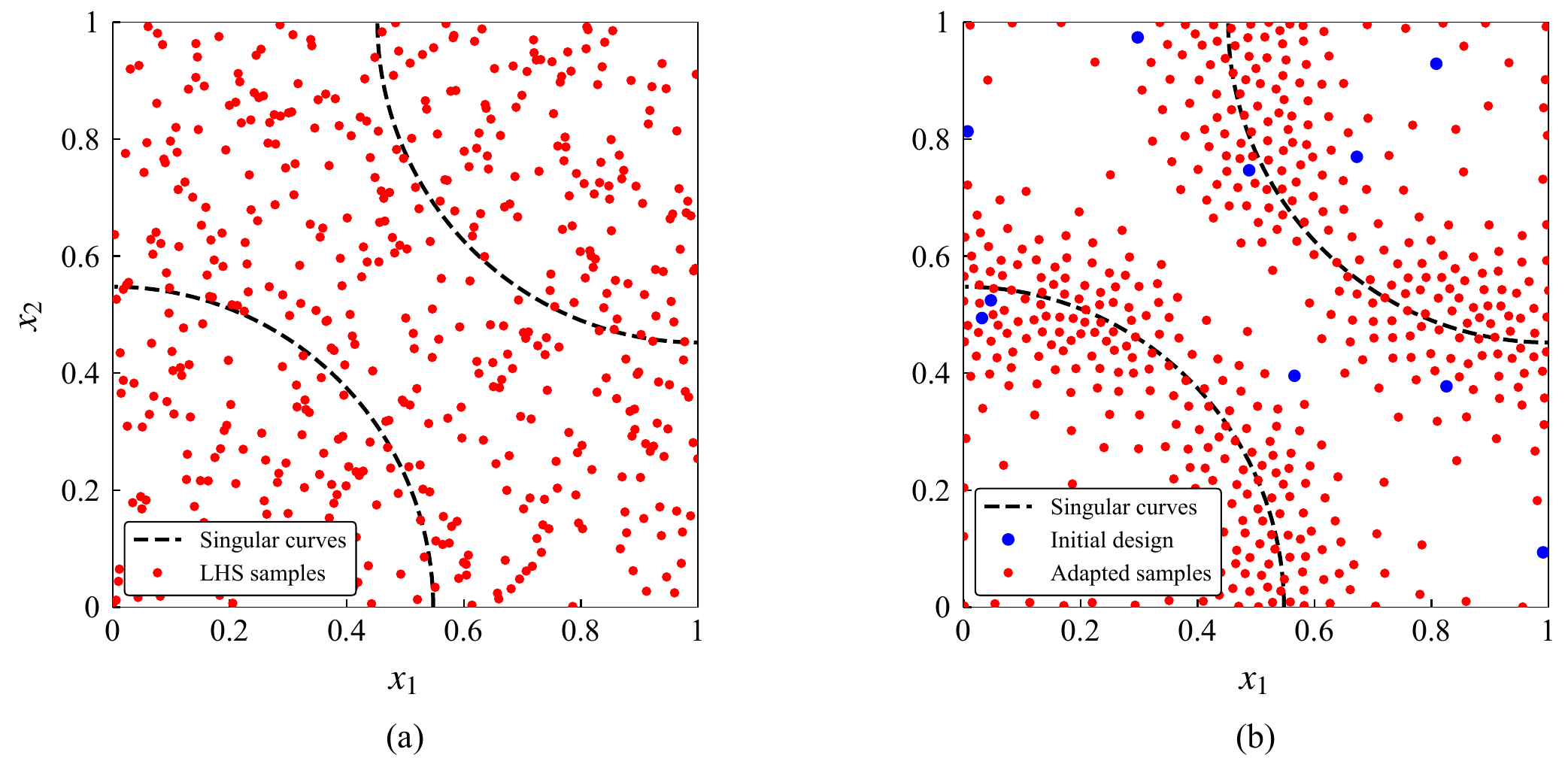}
    \caption{Sample distributions for the 2D mirror-line singularity problem: (a) samples generated by LHS and (b) initial and $\Theta$-adapted samples in the input space.}
    \label{fig:example1_1}
\end{figure}

\begin{figure}[h]
    \centering
    \includegraphics[width=1\linewidth]{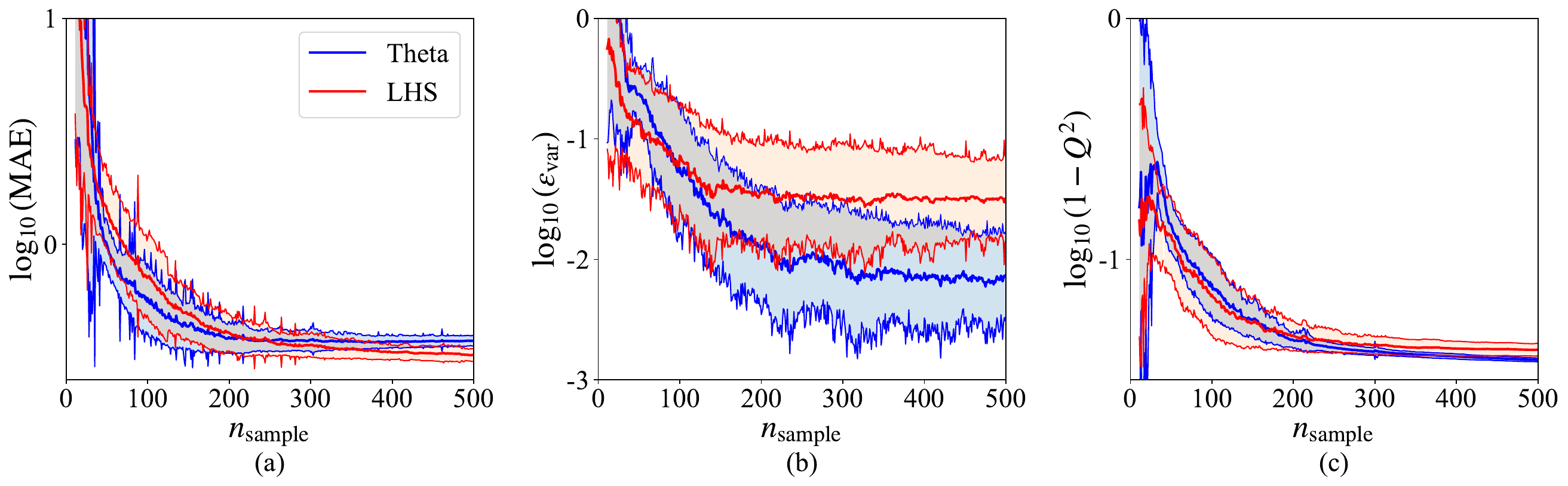}
    \caption{2D mirror line singularities: comparison of $\Theta$ criterion and LHS for PCE surrogate model construction: mean and standard deviation of (a) MAE, (b) relative variance error, and (c) LOO error. }
    \label{fig:example1_2}
\end{figure}

As shown in Fig.~\ref{fig:example1_1} (b), in a representative run, the training points selected by $\Theta$ criterion are primarily concentrated near the circular singularity curves. In contrast to the uniform distribution of the space-filling LHS design in Fig.~\ref{fig:example1_1} (a), the $\Theta$-adapted samples exhibit a clear clustering behavior along regions with a steep variance. Based on these samples, Fig.~\ref{fig:example1_2} presents a comparison of the mean and standard deviation of the MAE, relative variance error, and LOO error over 100 independent runs in the log-transformed space. Figure~\ref{fig:example1_2} (a) shows that, when the number of training samples is small, the $\Theta$ criterion reduces the MAE more rapidly and reaches a plateau with fewer samples compared to LHS. Although a slight increase in MAE is observed as the number of training samples continues to grow, the $\Theta$-based approach maintains a consistently narrower standard deviation band and less sensitive to the randomness of repeated trials. A similar convergence trend can be observed in Fig.~\ref{fig:example1_2}(b) for the relative variance error. The $\Theta$-based design exhibits a faster decay and converges to a noticeably lower error, while maintaining variability comparable to that of LHS in the log-scale. This suggests that adaptive sampling in high-variance regions enhances the capability of the PCE surrogate to capture second-order statistics more accurately. For LOO error, Fig.~\ref{fig:example1_2}(c) shows that the $\Theta$ criterion decreases slightly faster than LHS and ultimately achieves a lower error level. The standard deviation band of the $\Theta$ criterion is significantly narrow compared to LHS, and the influence of different realizations on the LOO error becomes negligible as the sample increases. This reflects the stronger robustness of the $\Theta$-adapted sampling strategy against variability across repeated realizations. Overall, the results demonstrate that the $\Theta$ criterion improves both accuracy and robustness of the PCE surrogate model with only marginal additional computational cost in this case.

\subsection{Reinforced Concrete Beam Problem}
\begin{figure}[h!]
    \centering
    \includegraphics[width=0.9\linewidth]{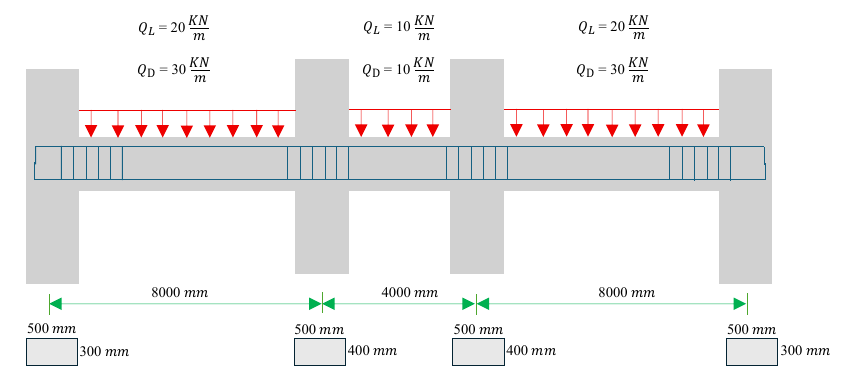}
    \caption{Geometry and loading configuration of the three-span reinforced concrete beam}
    \label{fig: example2_1}
\end{figure}

Reinforced concrete structures require rigorous design optimization to satisfy numerous code-based constraints safely and efficiently \cite{zhang2023influence}. However, evaluating these constraints often involves complex finite element-based simulations, which are computationally demanding—especially within iterative optimization frameworks. To demonstrate the effectiveness of the proposed sampling criterion for PCE surrogate model, a three-span continuous reinforced concrete beam is analyzed as the second example (see Fig.~\ref{fig: example2_1}), with parameters adopted from \cite{ozturk2024research}. Due to the statically indeterminate nature of the beam, structural responses must be evaluated at multiple critical sections to determine the required reinforcement and dimensions. Here, the PCE surrogate model is trained to estimate 15 primary structural responses: the bending moments at three locations per span (both ends and mid-span), totaling 9 values, and the shear forces at both ends of each span, totaling 6 values. These outputs are influenced by eight random variables, including dead and live loads, as well as material properties $f_{cd}$ and $f_{yd}$, whose statistical characteristics are summarized in Table~\ref{tab: example2_1}. Finally, the design constraints are formulated based on the flexural and shear capacities as presented in Equations~(17a) and~(17b).

\begin{subequations}
\label{eq:design_constraints}
\begin{align}
    M_{r}(i) &= A_s f_{yd} \left( d - \frac{A_s f_{yd}}{2 b_w f_{cd}} \right) \geq M_d(i) \label{eq:flexural_constraint} \\
    V_{r}(j) &= \frac{A_{sw}}{s} f_{ywd} d \geq V_d(j) \label{eq:shear_constraint}
\end{align}
\end{subequations}
where \(M_d(i)\) and \(M_r(i)\) represent the design bending moment and moment resistance at location \(i\) (including $n$ supports and $k$ mid-spans), while \(V_d(j)\) and \(V_r(j)\) denote the design shear force and shear resistance at the ends of each span.

\begin{figure}[h]
    \centering
    \includegraphics[width=1\linewidth]{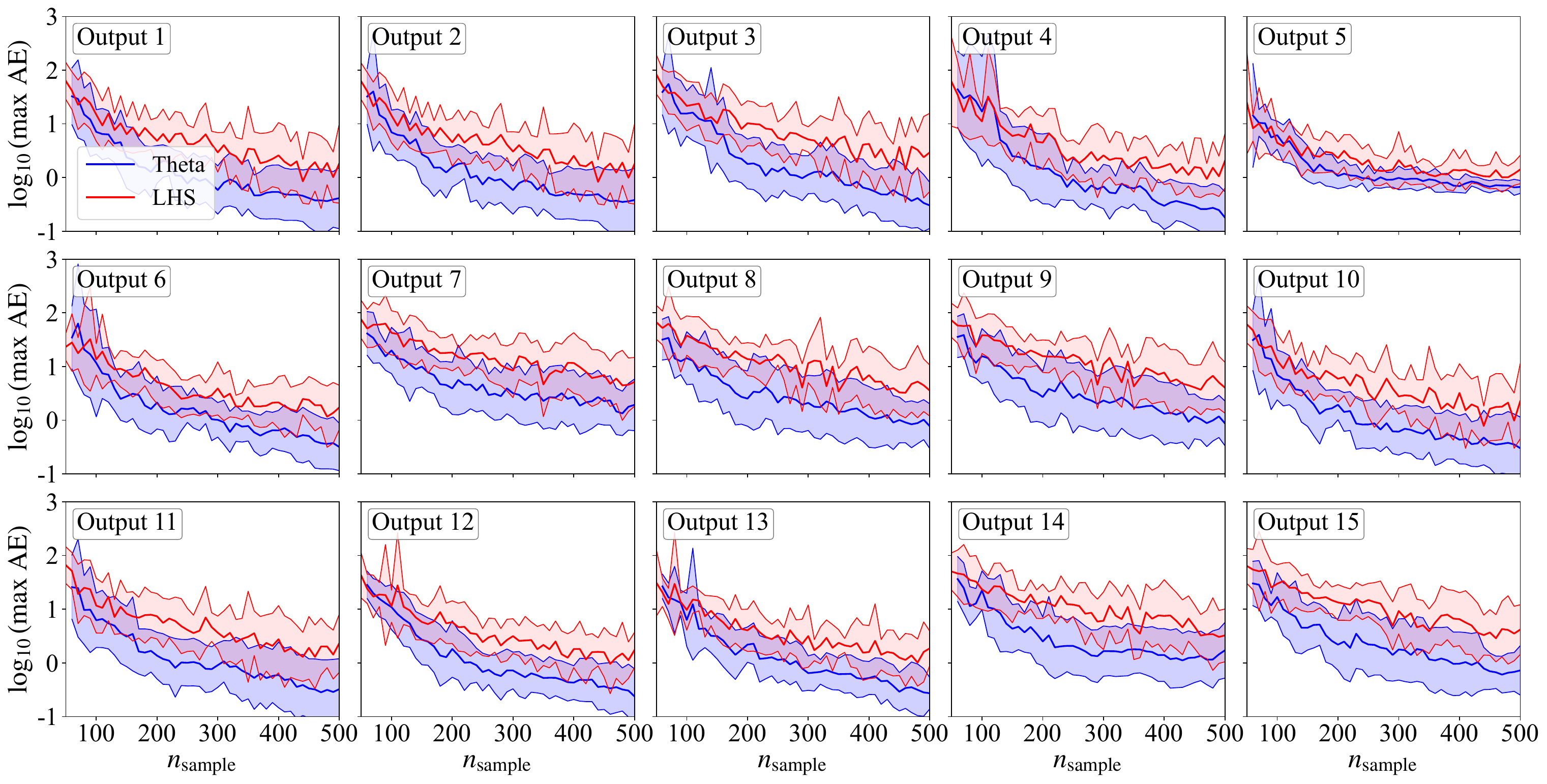}
    \caption{Reinforced concrete beam: comparison of $\Theta$ criterion and LHS for PCE surrogate model construction: mean and standard deviation of maximum absolute error.}
    \label{fig:example2_2}
\end{figure}

\begin{figure}
    \centering
    \includegraphics[width=1\linewidth]{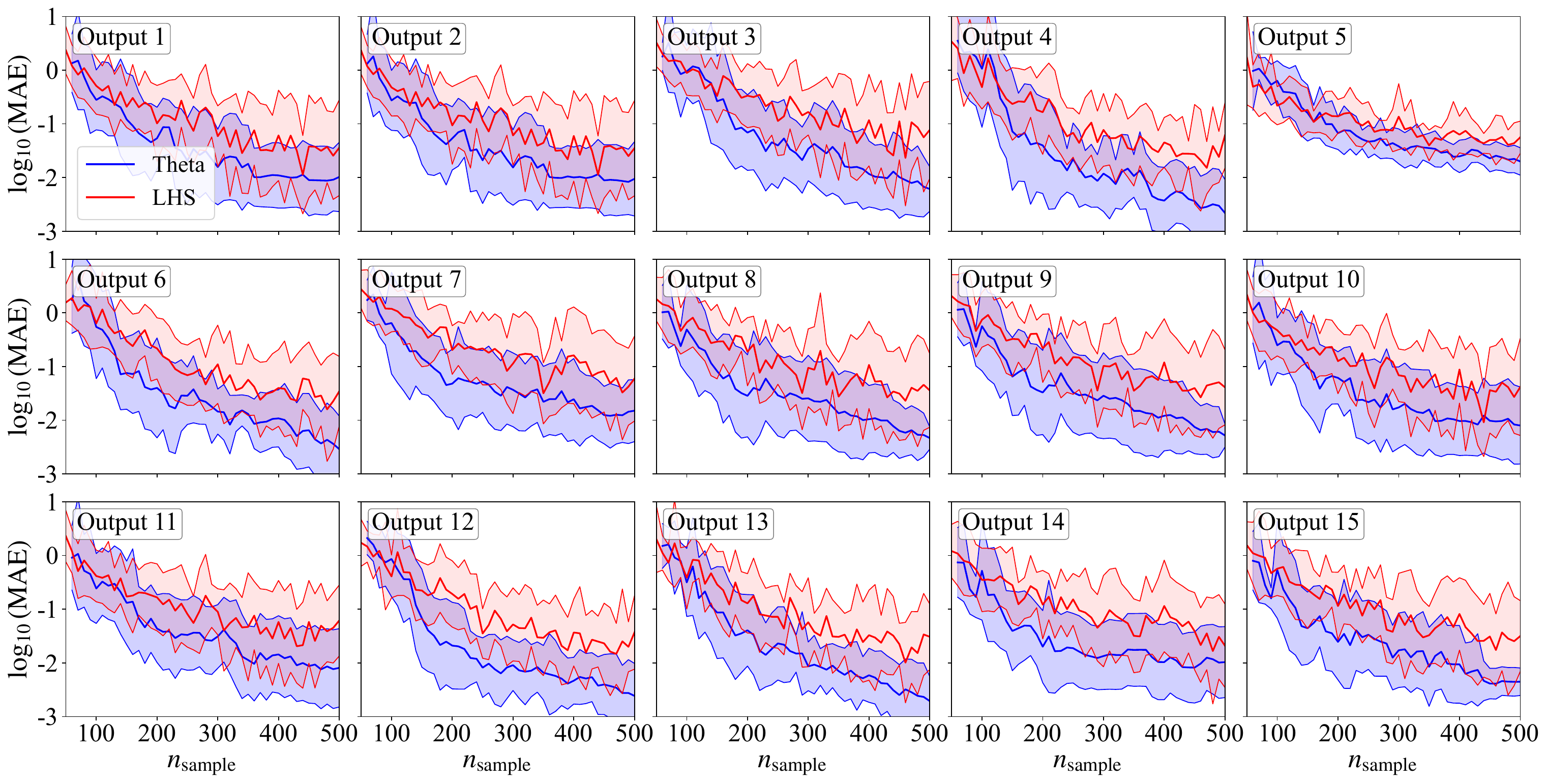}
    \caption{Reinforced concrete beam: comparison of $\Theta$ criterion and LHS for PCE surrogate model construction: mean and standard deviation of MAE.}
    \label{fig:example2_3}
\end{figure}

\begin{figure}
    \centering
    \includegraphics[width=1\linewidth]{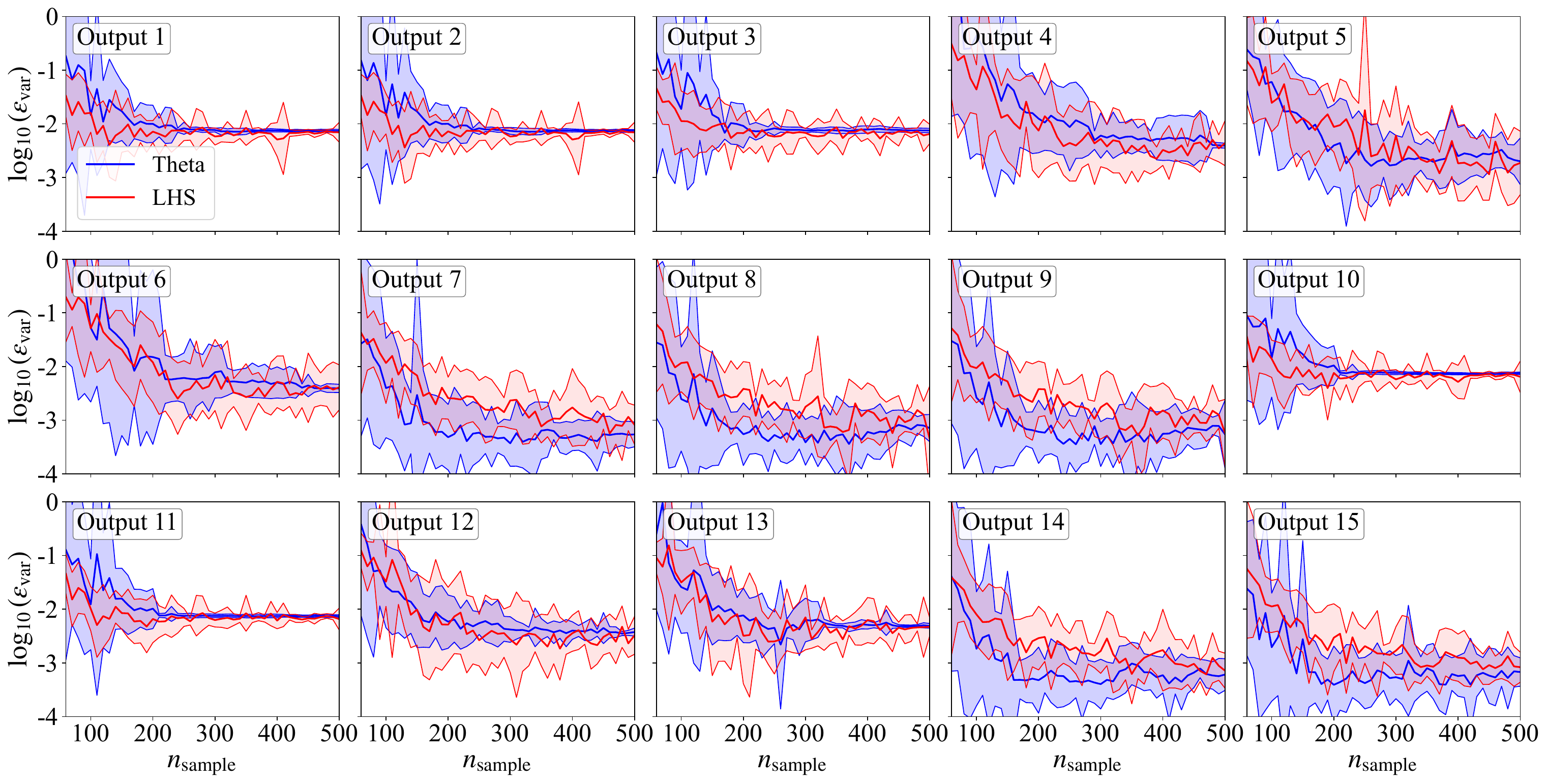}
    \caption{Reinforced concrete beam: comparison of $\Theta$ criterion and LHS for PCE surrogate model construction: mean and standard deviation of relative variance error.}
    \label{fig:example2_4}
\end{figure}

\begin{figure}
    \centering
    \includegraphics[width=1\linewidth]{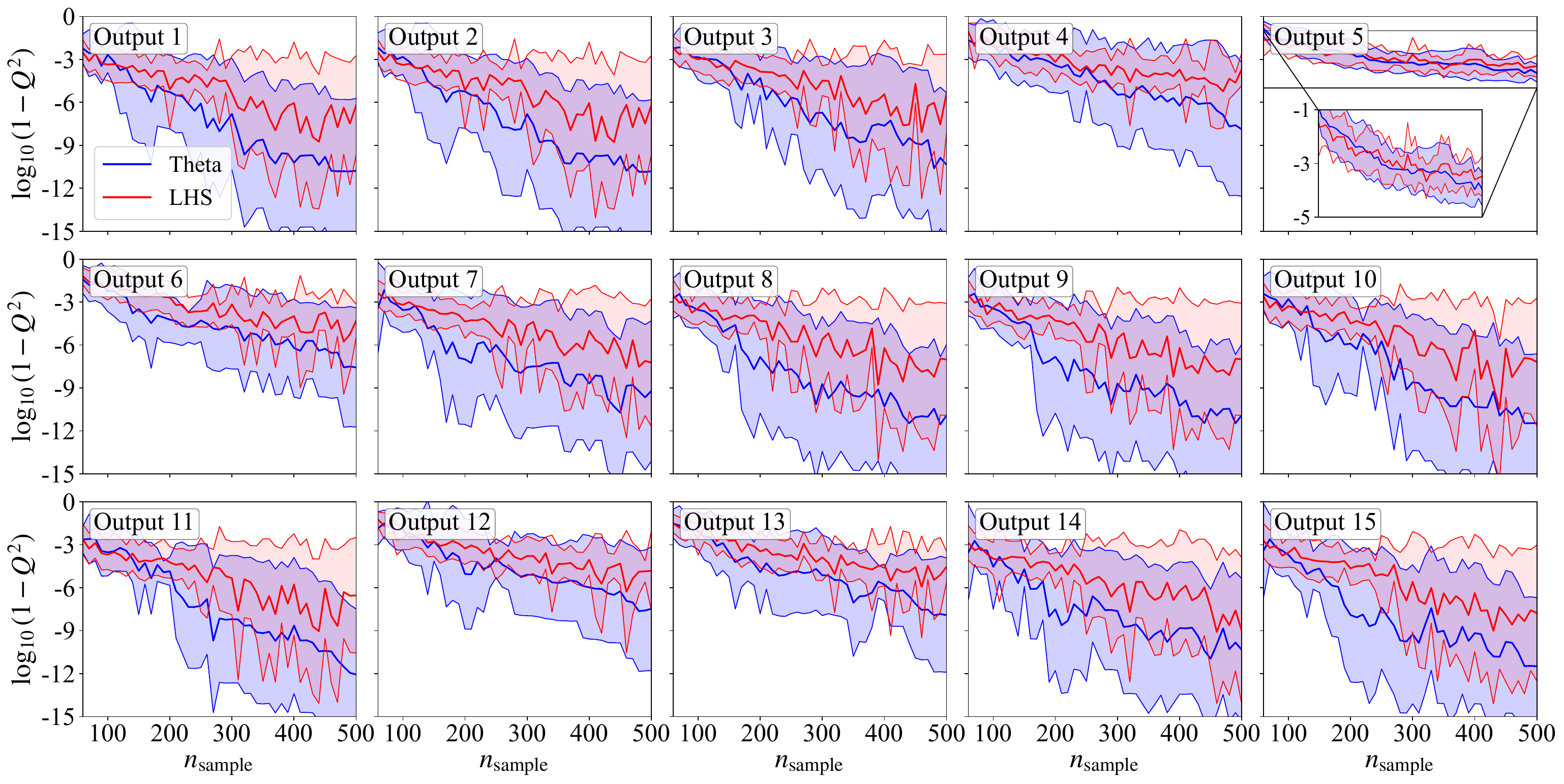}
    \caption{Reinforced concrete beam: comparison of $\Theta$ criterion and LHS for PCE surrogate model construction: mean and standard deviation of LOO error.}
    \label{fig:example2_5}
\end{figure}

In addition to the error metrics used previously, we further introduce the $\mathrm{AE}_{\max}$ in this example to provide a more comprehensive evaluation, which is particularly relevant in engineering applications, as it represents the worst-case prediction error. As shown in Fig. \ref{fig:example2_3}, when the initial surrogate models of two sampling strategies are at comparable levels of MAE, the $\Theta$ criterion rapidly exhibits a pronounced advantage as $n_{\mathrm{sample}}$ increases. The results in terms of the $\mathrm{AE}_{\max}$ exhibit similar trends (in Fig.~\ref{fig:example2_2}) indicating that the worst-case prediction error is also effectively reduced. In terms of the relative variance error, for outputs 7, 8, 9, 14, and 15, the $\Theta$ criterion demonstrates superior performance in the early stages of the sampling process, while approaching a performance comparable to LHS in the later stages. For the remaining outputs, the $\Theta$ criterion remains comparable to LHS throughout the entire sampling process (in Figure \ref{fig:example2_4}). A possible explanation for this behavior is that the normalization step in the $\Theta$ criterion reduces the relative contribution of low-variance regions within each output. Consequently, regions with dominant local variance tend to govern the sampling process, while other regions become less influential. Nevertheless, from a global perspective, this behavior is still acceptable, as the $\Theta$ criterion still shows a generally superior performance. Moreover, the $\Theta$ criterion tends to produce narrower standard deviation regions in the later stages, indicating lower variability and improved robustness against repeats. For the stability of the surrogate model, the $\Theta$ criterion exhibits a clear advantage over LHS, as illustrated in Fig.~\ref{fig:example2_5}. As the number of samples increases, the $\Theta$ criterion rapidly outperforms LHS, and the gap continues to widen. Overall, this example confirm that the proposed $\Theta$ criterion provides a more effective sampling trategy for multi-output PCE modeling in terms of model accuracy,  stability, and aggregate performance of variance estimation.

\begin{table}[ht]
\centering
\caption{Reinforced concrete beam: statistical properties of random variables}
\label{tab: example2_1}
\begin{tabular}{@{}lllll@{}}
\toprule
Variable & Symbol & Distribution & Mean & CoV (\%) \\
\midrule
\multirow{3}{*}{Dead load (KN/m)} 
    & $X_1$ & Normal & 30 & 10 \\
    & $X_2$ & Normal & 10 & 10 \\
    & $X_3$ & Normal & 30 & 10 \\
\midrule
\multirow{3}{*}{Live load (KN/m)}
    & $X_4$ & Gumbel & 20 & 25 \\
    & $X_5$ & Gumbel & 10 & 25 \\
    & $X_6$ & Gumbel & 20 & 25 \\
\midrule
Reinforcement yield strength (MPa) 
    & $X_7$ & Normal & 435 & 10 \\
\midrule
Concrete strength (MPa) 
    & $X_8$ & Log-normal & 1.46 & 15 \\
\bottomrule
\end{tabular}
\end{table}

\subsection{Offshore Wind Turbine Problem}
Offshore wind turbines (OWTs) operate under complex environmental conditions, where structural responses are influenced by various external factors such as wind and wave. The interaction between these environmental loads and the structural system can lead to nonlinear and coupled behavior, which poses challenges to accurate response prediction and reliability assessment \cite{FANG2026100035,ZHANG2025113309}. In this section, two representative cases are considered. The first case focuses on wind-induced structural responses, where the system is driven by a single environmental input, allowing the application of high polynomial orders for the surrogate without incurring excessive computational cost. The second case addresses a more realistic condition involving multiple environmental inputs and their coupled effects on the structural response. To this end, the NREL 5 MW monopile OWT from the OC3 project \cite{jonkman2009definition} is adopted as the benchmark structure for both cases.

\subsubsection{Offshore wind turbine moment responses under wind excitation}
This case investigates the structural response of an OWT under stochastic wind loading. The mean wind speed is treated as a random input following a Weibull distribution with a mean value of 12 m/s and a coefficient of variation (CoV) of 16.7\%, whereas the associated turbulent wind field leads to variability in the response \cite{sehrawat2016dynamic,FANG2026100035}. The response contains two bending moments, namely the edgewise and flapwise moments, which are closely related to fatigue and ultimate load effects.

\begin{figure}
    \centering
    \includegraphics[width=1\linewidth]{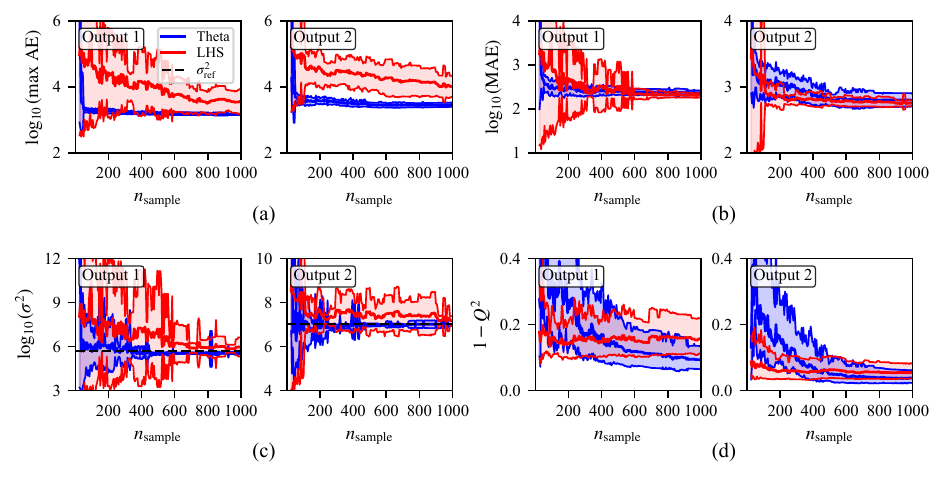}
    \caption{OWT under wind excitation: comparison of $\Theta$ criterion and LHS for PCE surrogate model construction: mean and standard deviation of maximum absolute error.}
    \label{fig:example3_1}
\end{figure}

Similarly to the previous example, four metrics (i.e., $\mathrm{AE}_{\max}$, MAE, variance, and LOO error) are considered to evaluate the performance of the $\Theta$ criterion. Fig.~\ref{fig:example3_1} (a) presents the evolution of the $\mathrm{AE}_{\max}$ for the two moments. The $\Theta$ criterion consistently achieves a lower $\mathrm{AE}_{\max}$ than LHS throughout the sampling process. Although $\mathrm{AE}_{\max}$ under LHS decreases as $n_{\mathrm{sample}}$ increases, it remains consistently worse. In contrast, the $\Theta$ criterion effectively controls the worst-case prediction error with only a limited number of additional samples. For MAE (Fig.~\ref{fig:example3_1} (b)), the LHS method exhibits substantial variability at small sample sizes, particularly for the first output. As the sample size increases, the uncertainty decreases and both methods converge toward stable error levels. For the first output, the two methods achieve comparable accuracy, whereas for the second output, LHS consistently yields a slightly lower MAE. For variance, the reference value $\sigma^2_{ref}$ is estimated using $1\times10^5$ Monte Carlo samples, which serves as the benchmark to assess the convergence of both sampling strategies. As shown in Fig.~\ref{fig:example3_1} (c), both methods eventually converge to $\sigma^2_{\mathrm{ref}}$. However, the $\Theta$ criterion requires only a small number of training samples to rapidly improve the accuracy and stability of the variance prediction, and consistently outperforms LHS throughout the sampling process. In terms of the stability of the surrogate, Fig.~\ref{fig:example3_1} (d) shows that the $\Theta$ criterion exhibits relatively higher values of $1 - Q^2$ in the early stages. When $n_{\mathrm{sample}}$ increases, the stability improves progressively, eventually becoming better than that of LHS. These observations collectively reflect the intrinsic balance of the $\Theta$ criterion: in the early stages, candidate points located in remote regions with extreme local variance may represent aggressive sampling choices, potentially compromising global approximation accuracy and space-filling, while effectively targeting regions associated with extreme responses; however, once selected, these points contribute to improving the global coverage of the input space through the distance-based component, thereby enhancing the overall stability and convergence behavior of the surrogate model.

\subsubsection{Offshore wind turbine displacement responses under multiphysics coupling}
\begin{figure}
    \centering
    \includegraphics[width=1\linewidth]{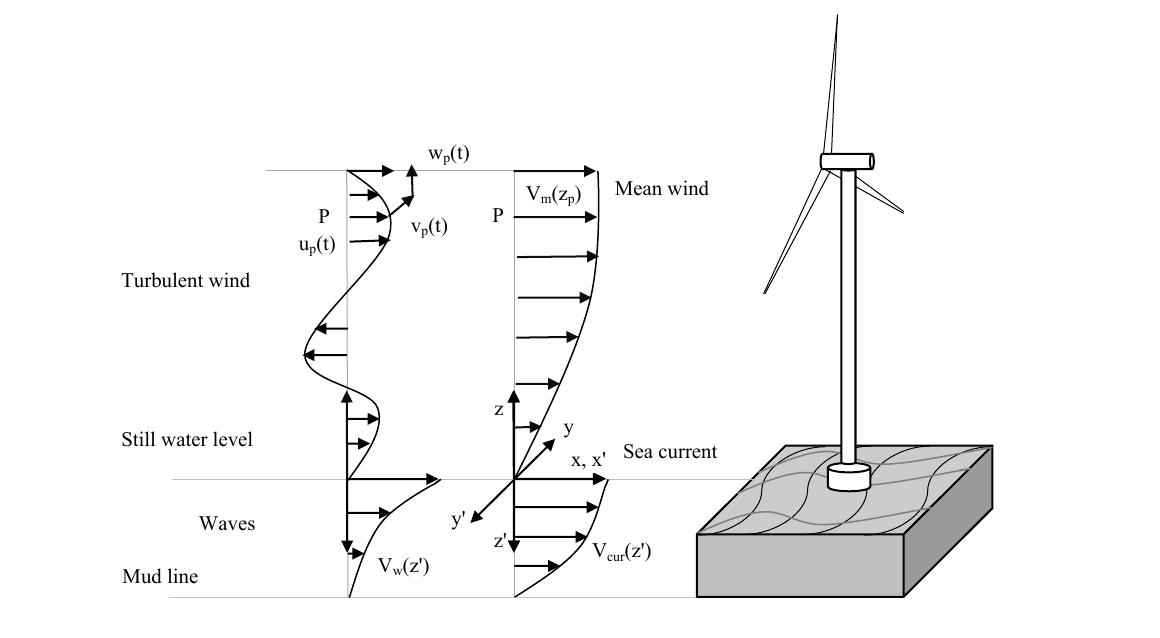}
    \caption{Schematic of multiphysics environmental interactions affecting the offshore wind turbine}
    \label{fig:example3_2_1}
\end{figure}

This case investigates the structural response of an OWT under coupled environmental loading conditions. Five environmental input variables are considered, including water depth, significant wave height, wind speed, wind direction, and wave direction. These variables jointly govern the coupled aerodynamic and hydrodynamic effects on the structure, leading to complex and nonlinear interactions within the system. The response is described by two displacement components in orthogonal directions, reflecting the global structural behavior under multiphysics coupling. Figure~\ref{fig:example3_2_1} provides a simplified representation of these environmental interactions affecting the OWT. The statistical properties of the environmental random variables are listed in Table~\ref{tab:environmental_vars} \cite{sehrawat2016dynamic}. Due to the increased number of input variables and the inherent nonlinearity of the responses, the number of basis functions in the PCE grows rapidly, resulting in a substantial computational cost of each simulation. It should be emphasized that this cost is dictated by the intrinsic complexity of the surrogate itself. The additional cost introduced by the $\Theta$ criterion is almost negligible. Therefore, instead of individual simulations for each realization, a set of 1000 samples is generated in advance using LHS, among which 100 selected samples are fixed as the testing set, while the remaining 900 samples are used as the initial training set and candidate pool for subsequent sampling. For comparison, the original LHS sampling is replaced by randomly selecting subsets from the LHS candidate pool in this case.

\begin{table}[t]
\centering
\caption{Multiphysics OWT: statistical properties of environmental random variables}
\label{tab:environmental_vars}
\begin{tabular}{@{}lllll@{}}
\toprule
Variable & Description & Distribution & Mean & CoV (\%) \\
\midrule
\( D_w \) & Water depth (m) & Normal & 25 & 10.0 \\
\( H_s \) & Significant wave height (m) & Weibull & 2 & 10.0 \\
\( U \) & Wind speed (m/s) & Weibull& 12 & 16.7 \\
\( \theta_{\text{wind}} \) & Wind direction (°) & Uniform & 90 & 10.0 \\
\( \theta_{\text{wave}} \) & Wave direction (°) & Uniform & 30 & 10.0 \\
\bottomrule
\end{tabular}
\end{table}

\begin{figure}
    \centering
    \includegraphics[width=1\linewidth]{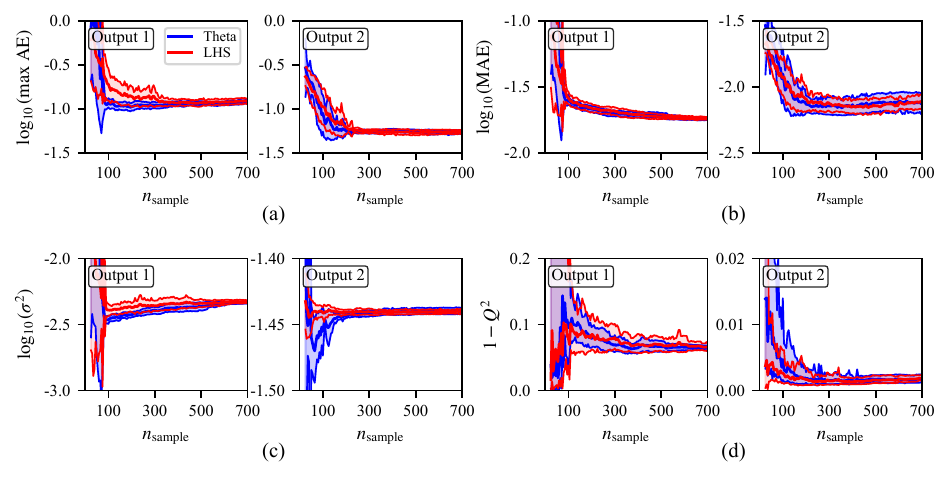}
    \caption{Multiphysics OWT: comparison of $\Theta$ criterion and LHS for PCE surrogate model construction: mean and standard deviation of maximum absolute error.}
    \label{fig:example3_2_2}
\end{figure}

The results of this case in Fig. \ref{fig:example3_2_2} exhibit similar trends to those observed in the previous case. The $\Theta$ criterion consistently achieves improved control of the worst-case prediction error, while maintaining comparable accuracy in terms of MAE. For variance, both sampling strategies converge towards a stable asymptotic value, with the $\Theta$ criterion showing a rapid decrease in the early stages. It should be noted that all the convergence behavior is influenced by the finite size of the candidate pool, which inherently limits the available sampling space and leads to similar asymptotic values for both methods towards the end of the sampling process. In particular, this effect is more evident for the $1 - Q^2$. Nevertheless, the $\Theta$ criterion exhibits a faster improvement in the early and intermediate stages, characterized by a more rapid decrease in $1 - Q^2$. Around $n_{\mathrm{sample}} = 200$, its performance becomes comparable to that of the LHS subset, and subsequently surpasses it by achieving lower values for output 1. Overall, the results from both cases demonstrate that the $\Theta$ criterion provides improved sampling efficiency across multiple outputs, particularly under limited sampling budgets. It enables more effective identification of regions associated with extreme responses, while also reducing variability in the surrogate predictions. Such characteristics are especially valuable for computationally demanding engineering applications, including reliability analysis and limit-state evaluation, where accurate characterization of extreme events is critical.

\section{Discussion \& Further Work}
\label{sec:discussion}
The results from the numerical examples shows that the proposed $\Theta$ criterion successfully improves sampling efficiency in multi-output problems. By aggregating variance contributions across multiple outputs, the method enables a joint utilization of multi-output information, thereby avoiding the computational cost and potential inconsistency caused by separate additional sampling. In addition, the multi-output $\Theta$ criterion exhibits reduced variability across different repeats, suggesting lower sensitivity to random sampling and improved robustness of the surrogate model. Despite these advantages, several limitations of the proposed approach should be acknowledged. First, the method relies on a predefined candidate pool, from which new samples are selected during the sequential enrichment process. In some high-dimensional engineering applications, such as the multiphysics OWT, constructing a sufficiently large and representative candidate pool generated by advanced sampling techniques \cite{VorMas:Nbody2:ADES:20} may become computationally prohibitive. As a result, the quality of the candidate pool may limit the achievable performance of the adaptive sampling strategy. Second, the computational cost associated with model updating may become significant in high-dimensional problems. In particular, when sparse regression techniques such as LAR are employed for PCE coefficient estimation, the iterative selection of basis functions can introduce additional computational overhead. While LAR has advantages for handling large candidate basis sets in high-dimensional cases, its cost may become non-negligible when the surrogate model is updated frequently during the sequential sampling process. These highlight the balance between sampling efficiency and increased computational cost, which should be considered in practical applications. A combination of the high computational burden of iterative sparse solvers with models containing large number of QoIs may be not feasible in practical applications. Possible solution can be seen in sparse solvers developed specifically for vector valued PCEs \cite{LOUKREZIS2025115746}.

A possible direction for future work is to further explore the correlation among multiple outputs. While the proposed approach already performs well, explicitly accounting for dependencies between outputs may be beneficial, especially when strong correlations are present. One way to achieve this is to employ covariance-based decomposition techniques, such as principal component analysis (PCA), to project the original outputs onto a set of orthogonal and uncorrelated components. The variance-based multi-output $\Theta$ criterion can then be formulated in this transformed space, where each component after projection represents an independent contribution to the overall variability. This transformation may help reduce redundancy among outputs and guide the sampling process toward the dominant modes. The results can also be mapped back to the original output space through an inverse transformation. Such an approach could be particularly useful for high-dimensional problems. The presented multi-output active learning can be also further combined with various existing techniques based on multiple PCE surrogates, such as multi-element PCEs \cite{MultielementPCE, NOVAK2023110728} for approximation of stochastic PDEs or other regression-based PCEs \cite{NOVAK2024112926,LOUKREZIS2025115746}.

\section{Conclusion}
This paper proposes a novel adaptive sequential sampling strategy for the construction of multi-output PCE surrogate models. The generalized $\Theta$ criterion extends the original single-output formulation by incorporating a normalized aggregation of variance contributions across multiple outputs, while maintaining a balance between variance-driven exploitation and distance-based exploration. The capability of the proposed approach has been demonstrated through several numerical examples, including both a synthetic function and engineering applications. The results consistently show that the $\Theta$ criterion improves surrogate accuracy and convergence behavior compared to LHS. In particular, it achieves more effective control of the worst-case prediction error, enhances the estimation of second-order statistics, and exhibits reduced variability across repeated runs, indicating improved robustness of the surrogate model. From an engineering perspective, the proposed method provides an efficient sampling strategy for multi-output systems under limited computational budgets. It enables more effective identification of regions associated with high variance or extreme responses, which is particularly beneficial for reliability analysis and UQ in complex systems. Overall, the proposed multi-output $\Theta$ criterion offers a practical and robust framework for adaptive sampling in PCE surrogate modeling and shows strong potential for applications in computationally demanding engineering problems.

\section*{Data Availability}
The data and scripts that support the findings of this study are openly available in https://github.com/NovakLBUT/MultTheta\_engineering at doi.org/10.5281/zenodo.20707012. 

\section*{Acknowledgments}

This work was supported by the Czech Science Foundation under project No. 26-20456S and project INODIN
(CZ.02.01.01/00/23\_020/0008487, Innovative methods of materials diagnostics and monitoring of
engineering infrastructure to increase its durability and service time) co-funded by European Union.

\bibliographystyle{elsarticle-num}
\bibliography{literatura}

\end{document}